# New HSL Distance Based Colour Clustering Algorithm

## Vasile Patrascu

Departement of Informatics Technology, Tarom Company
Calea Bucurestilor 224F, Bucharest, Romania,
e-mail: patrascu.v@gmail.com

## Abstract

In this paper, we define a distance for the HSL colour system. Next, the proposed distance is used for a fuzzy colour clustering algorithm construction. The presented algorithm is related to the well-known fuzzy c-means algorithm. Finally, the clustering algorithm is used as colour reduction method. The obtained experimental results are presented to demonstrate the effectiveness of our approach.

## Introduction

A colour image generally contains tens of thousands of colours. Therefore, most colour image processing applications first need to apply a colour reduction method before performing further sophisticated analysis operations such as segmentation. The using of colour clustering algorithm could be a good alternative for colour reduction method construction. In the framework of colour clustering procedure, we are faced with two colour comparison subject. We want to know how similar or how different two colours are. In order to do this comparison, we need to have a good coordinate system for colour representation and also, we need to define an efficient inter-colour distance measure in the considered system.

In this paper we will consider the particular case of colour representation by the perceptual system $HSL$ (hue, saturation and luminosity) (Smith 1978). The obtained degree of similarity or dissimilarity is dependent on the used inter-colour distance (Carron 1995), (Lim et al. 1990), (Sarifuddin et al. 2005), (Trivedi et al. 1986). In the $HSL$ colour comparison procedure, the three colour components have not the same importance. Thus, the most important is the hue, the saturation is the next and finally, the luminosity is less important. Because of this reason, we can define some colour clustering methods that use the bi-dimensional space $HS$. On this way, it is neglected the third component, the luminosity. This method is very good for those images that are quite saturated. In the same time, this method has the disadvantage of supplying erroneous clusters for the less saturated colours because it does not take into account the achromatic component, the luminosity. On the other side, when the luminosity is taken into account in the distance formula, there are arising situations when two colours are placed in different clusters because of the difference between their luminosities despite of their strong similarity in the chromatic space $HS$.

In this paper, we propose a distance that minimizes the two disadvantages mentioned above. This distance is constructed as a quasi-linear combination of the three standard distances of $H, S, L$ scalar components. Next, this distance is used for a fuzzy colour clustering construction.

In the following sections, the paper is thus organized: section 2 presents the particular form of the system $HSL$ used in this paper; section 3 presents the new colour distance; section 4 presents the colour clustering algorithm based on the proposed distance and related to the fuzzy c-means algorithm; section 5 presents some experimental results while section 6 outlines the conclusions.

## The Perceptual Colour System HSL

The most part of the colour images are represented by the $RGB$ colour system. However, this space presents the following two important limitations: the difficulty to determine colour features like the presence or the absence of a given colour, and the inability of the Euclidean distance to correctly capture colour differences in the $RGB$ space. Starting from the $RGB$ system, there were defined other systems for colour representation. One of them is the $HSL$ system where $H$ is the hue, $S$ is the saturation and $L$ is the luminosity. Colour space $HSL$ is also commonly used in image processing. As opposed to $RGB$ system, $HSL$ is considered as natural representation colour system. The $HSL$ system belongs to the perceptual system category because it is very close to the human colour perception. In the $HSL$ system, colour is decomposed according to physiological criteria like hue, saturation and luminosity. Hue refers to the pure spectrum colours and corresponds to dominant colour as perceived by human. Saturation corresponds to the relative purity or the quantity of white light that is mixed with hue while luminosity refers to the amount of light in a colour (Gonzales et al. 2007). A great advantage of *HSL* system over the *RGB* lies in its capacity to recognize the presence of colours in a given image.
There exist many formulae for $H, S, L$ components calculation. For the hue $H$, the most part of the definitions are closed to the following formula that uses the *atan2* function (ECMA-262 2011):

$$H = \operatorname{atan2}\left(\frac{B-G}{\sqrt{2}}, \frac{2R-B-G}{\sqrt{6}}\right) \quad (1)$$

and $H \in (-\pi, \pi]$.
For the luminosity calculation, in this paper, the following formula will be used (Patrascu 2012):

$$L = \frac{M}{1+M-m} \quad (2)$$

where

$$M = \max(R, G, B)$$

and

$$m = \min(R, G, B).$$

Regarding to the saturation calculation, many variants are defined by the distance between $\max(R, G, B)$ and $\min(R, G, B)$. In this paper, we will use for saturation calculation, the following formula (Patrascu 2009):

$$S = \frac{2(M-m)}{1+|M-0.5|+|m-0.5|} \quad (3)$$

We suppose $R, G, B \in [0,1]$ and it results that $M, m, S, L \in [0,1]$.

## The New HSL Colour Distance

In the Cartesian coordinate system $(x, y, z) \in R^3$, for two vectors $v_1 = (x_1, y_1, z_1)$, $v_2 = (x_2, y_2, z_2)$ one defines the Euclidean distance by:

$$D_E^2(v_1, v_2) = (x_1 - x_2)^2 + (y_1 - y_2)^2 + (z_1 - z_2)^2$$

In order to obtain the variant of Euclidean distance for the cylindrical coordinate system, we use the following substitution:

$$x = \rho \cos(\varphi), \ y = \rho \sin(\varphi), \ z = l.$$

It results:

$$D_E^2(v_1, v_2) = 4\rho_1 \rho_2 \cdot d^2(\varphi_1, \varphi_2) + d^2(\rho_1, \rho_2) + d^2(l_1, l_2)$$

were:

$$d^2(\varphi_1, \varphi_2) = \sin^2\left(\frac{\varphi_1 - \varphi_2}{2}\right)$$

$$d^2(\rho_1, \rho_2) = (\rho_1 - \rho_2)^2$$

$$d^2(l_1, l_2) = (l_1 - l_2)^2$$

The coordinate system *HSL* is a cylindrical one and for two colours $Q_1 = (H_1, S_1, L_1)$ and $Q_2 = (H_2, S_2, L_2)$, the Euclidean distance becomes (Gonzales et al. 2007):

$$D_E^2(Q_1, Q_2) = 4S_1 S_2 \cdot d^2(H_1, H_2) + d^2(S_1, S_2) + \\ + d^2(L_1, L_2) \quad (4)$$

The colour Euclidean distance $D_E$ has three terms: the distance between hues, the distance between saturations and the distance between luminosities. The distance between hues is multiplied by a factor that depends on the colour saturations. This factor has a multiplicative structure. Thus, when the saturation values increase, the hues distance influence increases in framework of distance $D_E$. When the saturation values decrease, the hue distance influence decreases.
From here, the idea to multiply the luminosity distance with a similar factor comes up. This factor will have the following behaviour: when the saturation values increase, the luminosity distance influence decreases and when the saturation values decrease, the luminosity distance influence increases.
We can generalize (4) using two real and positive parameters $\alpha$ and $\beta$, by the following:

$$D_P^2(Q_1, Q_2) = \alpha \cdot d^2(H_1, H_2) + \beta \cdot d^2(L_1, L_2) + \\ + d^2(S_1, S_2) \quad (5)$$

The parameter $\alpha$ will be related to the colour chromaticity while the parameter $\beta$ will be related to the colour achromaticity. Before the construction of parameters $\alpha$ and $\beta$, we define the index of chromaticity $c$ and the index of achromaticity $a$, having the following properties:

- index of chromaticity
$c : [0,1] \rightarrow [0,1]$,
    (i) $c(0) = 0$
    (ii) $c(1) = 1$
    (iii) $\forall S_1, S_2 \in [0,1]$,
        if $S_1 < S_2$ then $c(S_1) < c(S_2)$.

We will use for the chromaticity index the following particular function:

$$c(S) = \sqrt{S} \qquad (6)$$

- index of achromaticity

$a : [0,1] \to [0,1]$,

(i) $a(0) = 1$,

(ii) $a(1) = 0$.

(iii) $\forall S_1, S_2 \in [0,1]$,
if $S_1 < S_2$ then $a(S_1) > a(S_2)$.

We will use for the achromaticity index the following particular function:

$$a(S) = \sqrt{1-S} \qquad (7)$$

The two parameters $\alpha$ and $\beta$ will have a multiplicative structure. Thus, $\alpha$ will be a product of chromaticity indexes, while $\beta$ will be a product of achromaticity indexes. It results:

$$\alpha = c(S_1) \cdot c(S_2) \qquad (8)$$

$$\beta = a(S_1) \cdot a(S_2) \qquad (9)$$

Using (6), (7), (8) and (9) for the two multipliers $\alpha$ and $\beta$, it results the following particular forms:

$$\alpha = \sqrt{S_1} \cdot \sqrt{S_2} \qquad (10)$$

$$\beta = \sqrt{1-S_1} \cdot \sqrt{1-S_2} \qquad (11)$$

The distance (5) becomes:

$$D_P^2(Q_1, Q_2) = \sqrt{S_1 \cdot S_2} \sin^2\left(\frac{H_1 - H_2}{2}\right) + \\ + \sqrt{(1-S_1) \cdot (1-S_2)} \cdot (L_1 - L_2)^2 + (S_1 - S_2)^2 \qquad (12)$$

## The Colour Clustering Algorithm

Let there be $n$ colours $Q_1, Q_2, \ldots, Q_n$ that must be clustered into $k$ sets. Each cluster $j$ is characterized by the membership coefficients $w_{j1}, w_{j2}, \ldots, w_{jn}$ for the considered $n$ colours and the cluster center defined by the colour $q_j = (h_j, s_j, l_j)$. For colour clustering, we will construct an algorithm that is similar to the fuzzy c-means algorithm (Bezdek 1981).

The membership function $w_{ij}$ is calculated using formula (13), where the distance $D_P$ is calculated with formula (12).

$$\begin{cases} \forall i \in [1,n] \\ \forall j \in [1,k] \end{cases},$$

$$w_{ij} = \frac{1}{1 + \sum\limits_{\substack{m=1 \\ m \neq j}}^{k} \left( \frac{D_P(Q_i, q_j)}{D_P(Q_i, q_m)} \right)^{\frac{2}{\omega - 1}}} \qquad (13)$$

The functions $w_{ij}$ verify the condition of the partition of unity, namely:

$$\forall i \in [1,n], \qquad w_{i1} + w_{i2} + \ldots + w_{ik} = 1$$

The cluster center components $(h_j, s_j, l_j)$ are calculated with formulae (14), (15) and (16).

$\forall j \in [1,k]$,

$$h_j = \operatorname{atan2}\left( \frac{\sum\limits_{i=1}^{n} w_{ij}^{\omega} C_i \cdot \sin(H_i)}{\sum\limits_{i=1}^{n} w_{ij}^{\omega} C_i}, \frac{\sum\limits_{i=1}^{n} w_{ij}^{\omega} C_i \cdot \cos(H_i)}{\sum\limits_{i=1}^{n} w_{ij}^{\omega} C_i} \right) \qquad (14)$$

with $C_i = \sqrt{S_i}$.

$\forall j \in [1,k]$,

$$l_j = \frac{\sum\limits_{i=1}^{n} w_{ij}^{\omega} A_i \cdot L_i}{\sum\limits_{i=1}^{n} w_{ij}^{\omega} A_i} \qquad (15)$$

with $A_i = \sqrt{1-S_i}$.

$$\forall j \in [1,k], \qquad s_j = \frac{\sum\limits_{i=1}^{n} w_{ij}^{\omega} \cdot S_i}{\sum\limits_{i=1}^{n} w_{ij}^{\omega}} \qquad (16)$$

where $\omega$ is a fuzzification-defuzzification parameter and also, $\omega \in (1, 1.5)$.

## Experimental Results

We have applied this method to images: "red-green" (figure 1a), "flower" (figure 2a), "house" (figure 3a), "parrots" (figure 4a) and "bird" (figure 5a). The clustered images obtained using the new distance can be seen in figures 1b, 2b, 3b, 4b, 5b and those obtained using the Euclidean distance can be seen in figures 1c, 2c, 3c, 4c, 5c.

Figure 1a shows a synthetic image. Using the proposed distance, it was obtain the 3-colour representation shown in figure 1b while figure 1c shows the image obtained using the Euclidean distance. In the first case, the black region has a small area, while in the second case, it has a large one. This fact proves the influence of achromatic parameter that was used in the proposed distance.

In the case of image "flower", using the Euclidean distance it was obtained two clusters for background. These two clusters have the same hue but the luminosities are different (figure 2c). Using the proposed distance, one obtained only one cluster for the entire background (figure 2b).

For image "house", using the Euclidean distance, the white and bright blue colours were not separated (figure 3c). In the case of image "parrots", the yellow and red colours were not separated (figure 4c) and the clustering is different from that obtained using the proposed distance (figure 4b). For the image "bird", using the Euclidean distance, the orange and grey colours were not separated (figure 5c).

The experimental results illustrate that our distance performs well compared to Euclidean distance. In the same time, the obtained results show the helpfulness of using this novel inter-colour distance measure in colour clustering algorithms.

## Conclusions

In this paper one presents an enhancement of fuzzy c-means algorithm for the particular case of colour clustering. It was used the perceptual colour system $HSL$ for colour representation. The main step is represented by definition of a new distance in the $HSL$ colour space. In this construction, there were used two multipliers that make the balance between the hue weight and luminosity weight in the framework of this three-term colour distance. We can conclude that the new inter-colour distance "$D_P$" defined by (12), the two multipliers "$\alpha$" defined by (8) and "$\beta$" defined by (9), the particular form of index of chromaticity "$c$" defined by (6) and achromaticity "$a$" defined by (7) construct a new and useful framework for colour clustering procedures. The obtained experimental results were compared with those obtained by using the Euclidean distance. This comparison shows the efficiency of the proposed clustering algorithm using for the colour reduction method.

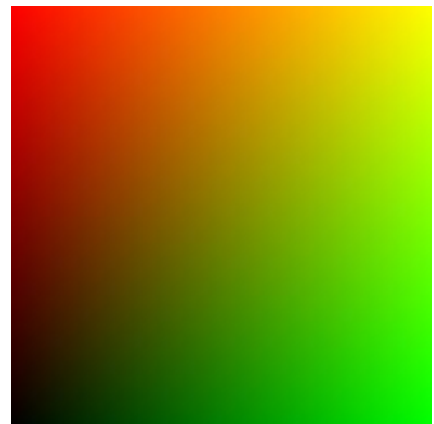
(a)

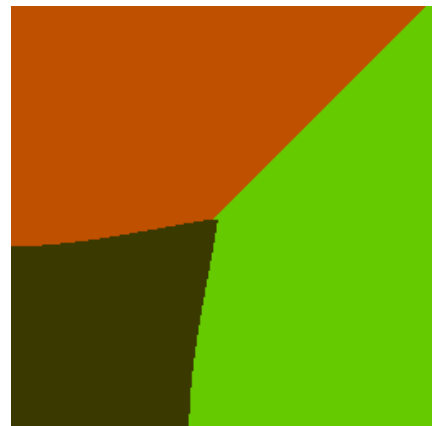
(b)

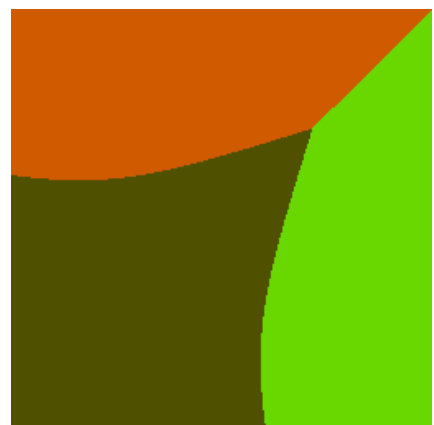
(c)

Figure 1: The image „red-green" (a) and its 3-colour representation based on proposed HSL distance (b) and based on HSL Euclidean distance (c).

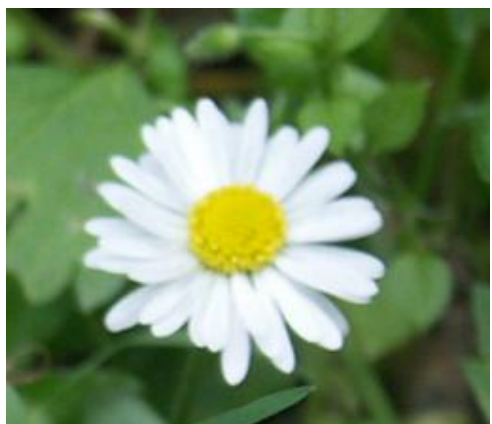
(a)

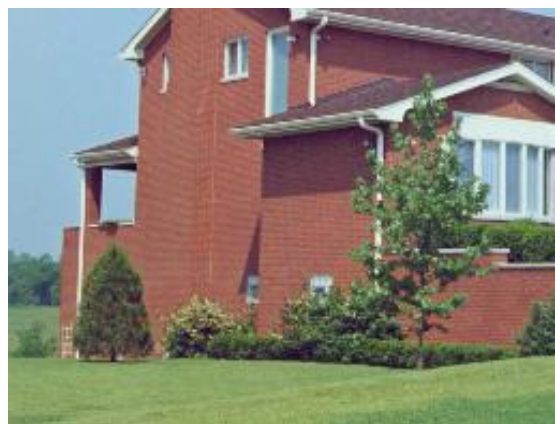
(a)

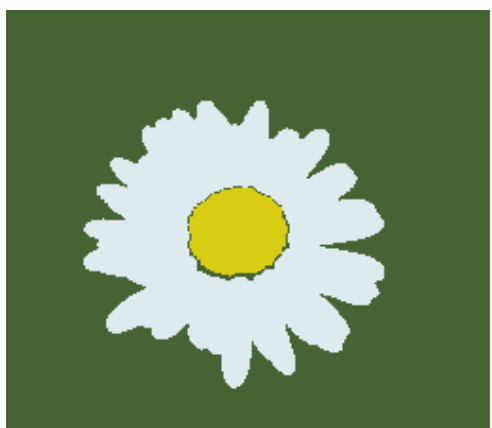
(b)

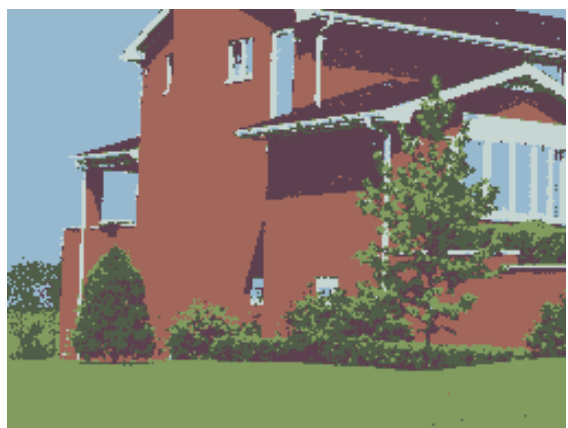
(b)

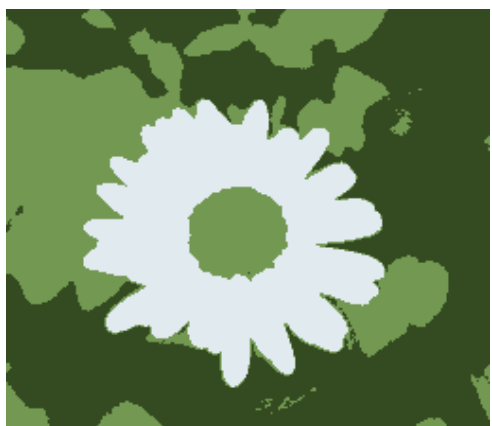
(c)

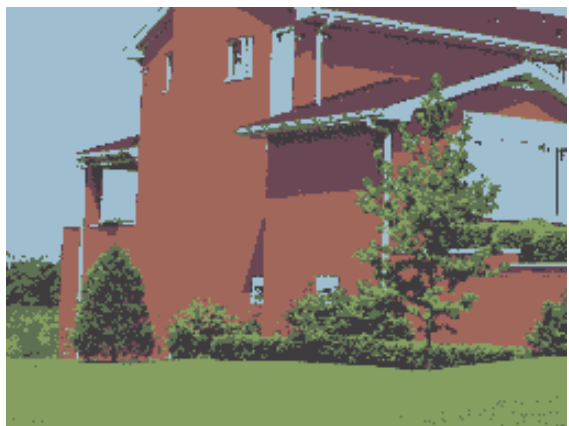
(c)

Figure 2: The image „flower" (a) and its 3-colour representation based on proposed HSL distance (b) and based on HSL Euclidean distance (c).

Figure 3: The image „house" (a) and its 6-colour representation based on proposed HSL distance (b) and based on HSL Euclidean distance (c).

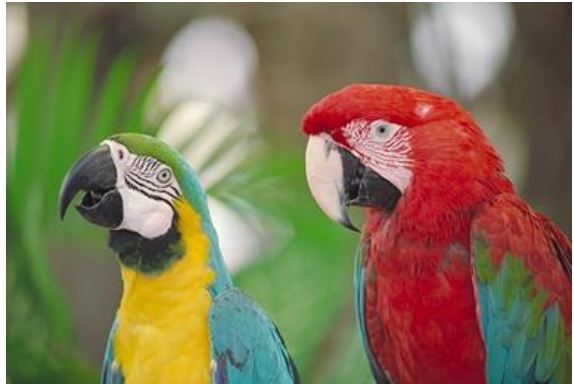

(a)

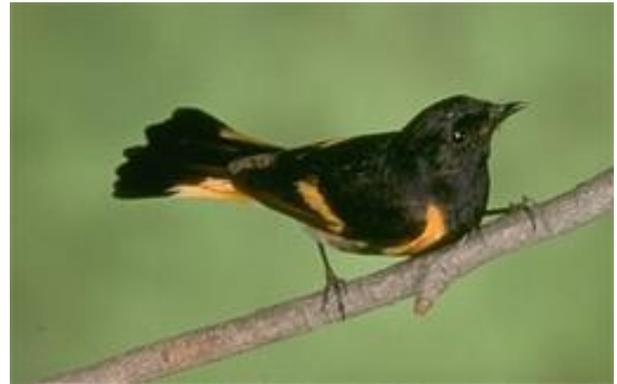

(a)

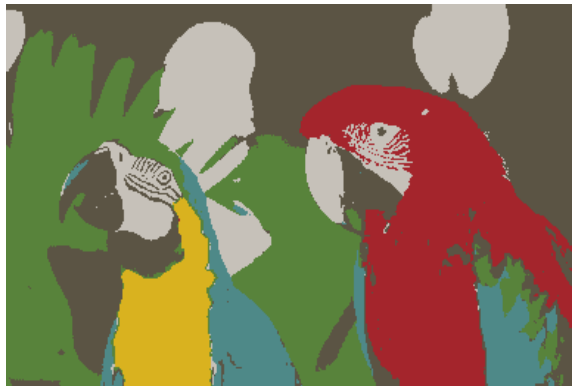

(b)

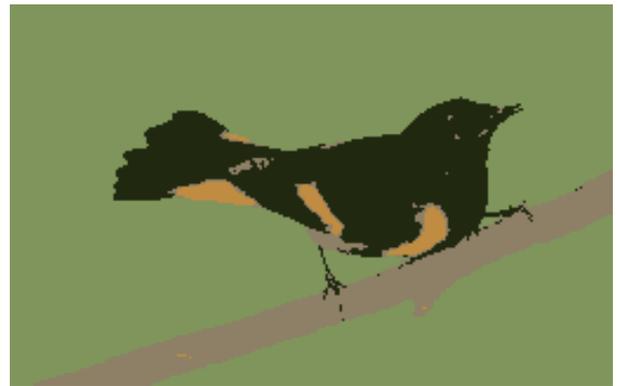

(b)

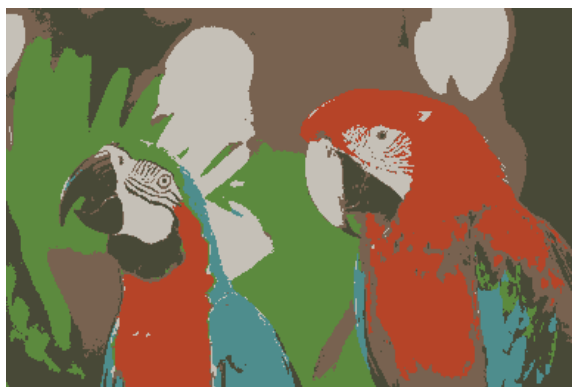

(c)

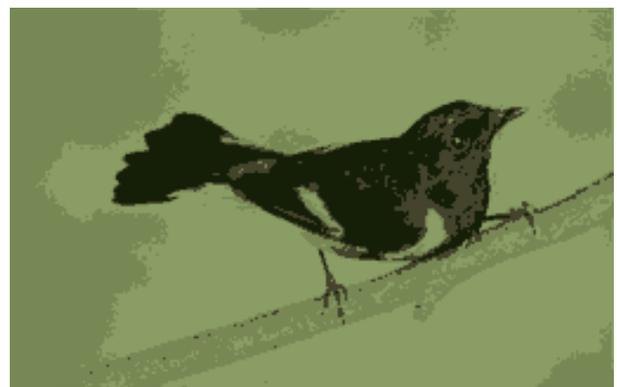

(c)

Figure 4: The image „parrots" (a) and its 6-colour representation based on proposed HSL distance (b) and based on HSL Euclidean distance (c).

Figure 5: The image „bird" (a) and its 4-colour representation based on proposed HSL distance (b) and based on HSL Euclidean distance (c).